\documentclass{article}

\usepackage{arxiv}

\usepackage[utf8]{inputenc} 
\usepackage[T1]{fontenc}    
\usepackage{hyperref}       
\usepackage{url}            
\usepackage{booktabs}       
\usepackage{amsfonts}       
\usepackage{nicefrac}       
\usepackage{microtype}      
\usepackage{lipsum}         
\usepackage{graphicx}
\usepackage{doi}
\usepackage{mathtools}
\usepackage[table,xcdraw]{xcolor}
\usepackage{multirow}
\usepackage{amsmath}
\usepackage{cleveref}       
\usepackage{bbm}
\usepackage{algorithm}
\usepackage{algorithmic}

\title{Hierarchical Decision Transformer}

\date{}

\author{ \hspace{1mm}André Correia \\
	NOVA LINCS\\
	Universidade da Beira Interior\\
	Covilhã, Portugal \\
	\texttt{andre.correia@ubi.pt} \\
	\And
	\hspace{1mm}Luís A. Alexandre \\
	NOVA LINCS\\
	Universidade da Beira Interior\\
	Covilhã, Portugal \\
	\texttt{lfbaa@di.ubi.pt} \\
}

\renewcommand{\shorttitle}{Hierarchical Decision Transformers}

\hypersetup{
pdftitle={Contrastive Learning from Demonstrations},
pdfsubject={q-bio.NC, q-bio.QM},
pdfauthor={André Correia, Luís A. Alexandre},
pdfkeywords={Contrastive, Learning, from, Demonstrations},
}

\begin{document}
\maketitle

\begin{abstract}
Sequence models in reinforcement learning require task knowledge to estimate the task policy.
This paper presents a hierarchical algorithm for learning a sequence model from demonstrations.
The high-level mechanism guides the low-level controller through the task by selecting sub-goals for the latter to reach. This sequence replaces the returns-to-go of previous methods, improving its performance overall, especially in tasks with longer episodes and scarcer rewards.
We validate our method in multiple tasks of OpenAIGym \cite{openai}, D4RL \cite{d4rl} and RoboMimic \cite{robomimic} benchmarks.
Our method outperforms the baselines in eight out of ten tasks of varied horizons and reward frequencies
without prior task knowledge, showing the advantages of the hierarchical model approach for learning from demonstrations using a sequence model.
\end{abstract}


\section{Introduction}
\label{introduction}

\begin{figure}[!tb]
  \centering
  \includegraphics[width=0.8\linewidth]{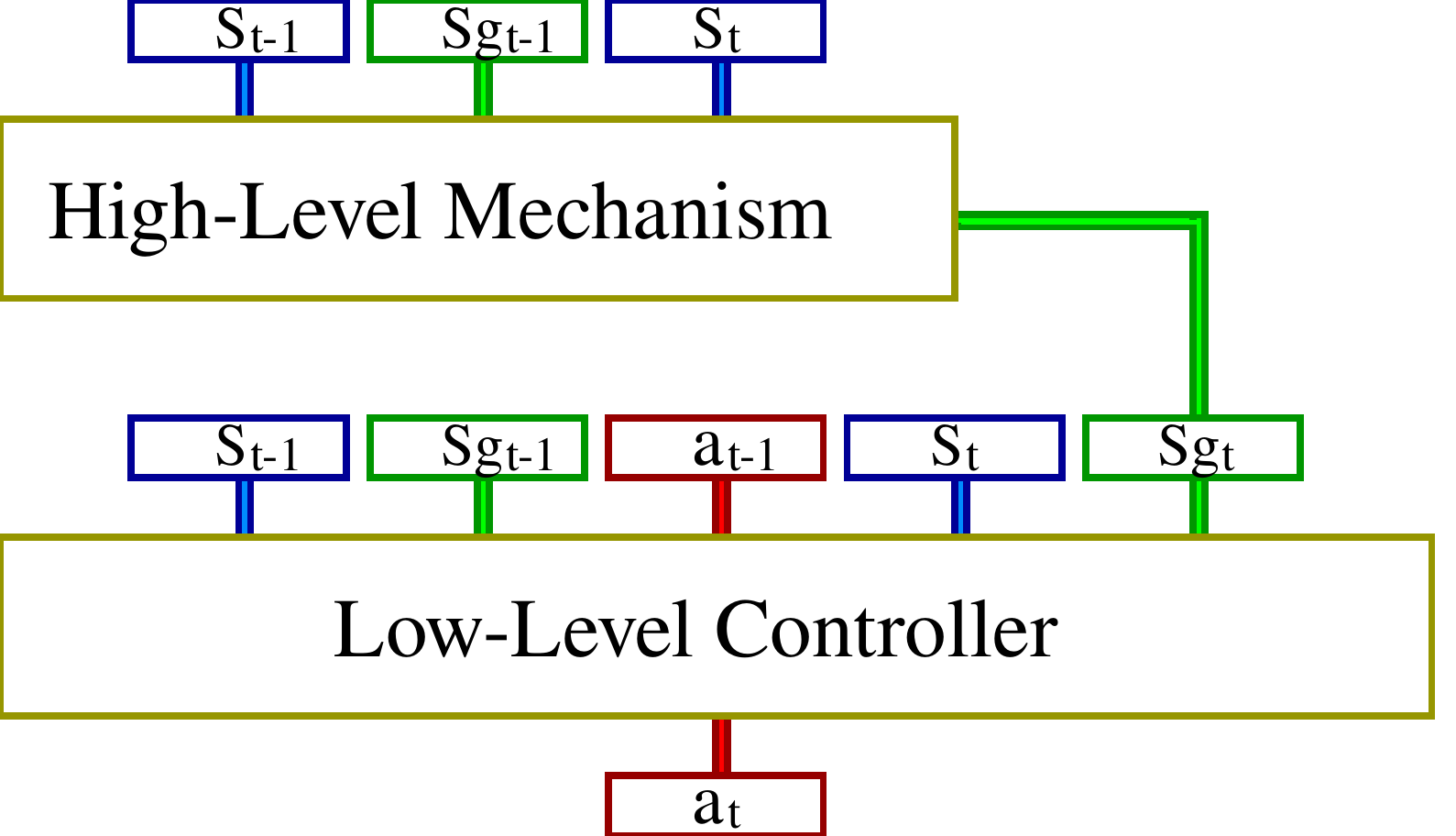} 
  
  \caption{HDT framework: We employ two decision transformer models in the form of a high-level mechanism and a low-level controller. The high-level mechanism guides the low-level controller through the task by selecting sub-goal states, based on the history of sub-goals and states, for the low-level controller to try to reach. The low-level controller is conditioned on the history of past states, sub-goals and actions to select the appropriate action. By reaching each sub-goal, the controller gets closer to completing the task. With this architecture, the model doesn't require a user to specify the initial value of desired rewards for a given task and becomes more robust to sparse rewards and tasks with long episodes.}
  \label{hdt}
\end{figure}

Reinforcement learning (RL) \cite{barto} has been successfully applied to a diverse set of robotic tasks, such as pushing and grasping objects, path finding and locomotion \cite{dreaming,stabilizing,conservative,ris}. 
However, RL algorithms require large amounts of exploration to successfully learn policies in large state and action spaces, which is costly in real-world applications.
Additionally, learning a policy through RL requires specifying a reward function that is specifically designed to assist in exploration. Furthermore, RL often relies on dense rewards, and struggles in environments with sparse rewards. Creating dense reward functions could mitigate such issues. However, it is often difficult to design such reward functions for complex real-world tasks.

We can alleviate the above-mentioned problems caused from learning policies from scratch by leveraging prior experience collected by an expert demonstrator as an extra supervision signal using Demonstration Learning (DL).
Because the demonstrator was optimizing a reward function, the data set can be used in lieu of a reward function to further alleviate some of the aforementioned issues.
When learning from demonstrations, the performance of the agent is dependent on the quality of the data set. However, due to the difficulty of collecting data, the majority of data sets are sub-optimal, unstructured and diverse. One way to utilize unstructured prior experience is to identify key states that contributed to the success of the trajectory. Hierarchical models are used to solve this issue, where a high-level model identifies sub-goals which the low-level model must reach, guiding it to the final goal and consequently solving the task.

Transformer models have revolutionized several machine learning fields and have recently been applied to RL in the form of decision transformers \cite{dt}. Here the traditional RL algorithms are replaced by a sequence modeling objective, avoiding some problems with the traditional algorithms. However, these models are dependent on frequent rewards to create the input sequences, limiting their applicability to short-horizon tasks. Furthermore, they require the specification of the accumulated desired rewards, which the user must carefully state for each task. The insight is to replace the reward sequence that guides the learning process with a sequence of key sub-goal states identified from the data set.

We propose the Hierarchical Decision Transformer, a dual decision transformer architecture which scales the original decision transformer to tasks with long episodes and removes the need for the specification of desired rewards by a user. The high-level transformer predicts key sub-goal states, identified from the demonstration data set, based on the sequence of prior states. The low-level transformer is the original decision transformer conditioned on the sequence of sub-goals instead of desired rewards.

\textbf{Summary of Contributions:}
\begin{enumerate}
    \item We present Hierarchical Decision Transformer (HDT), a dual transformer framework that enables offline learning from a large set of diverse and sub-optimal demonstrations by selectively selecting sub-goal states from the dataset.
    \item We evaluate HDT on various environments and data sets from the D4RL \cite{d4rl}, OpenAI Gym \cite{openai} and RoboMimic \cite{robomimic} benchmarks.
    
    \item We show that our method outperforms the original decision transformer baseline, especially in tasks with longer episodes, proving that the sub-goal sequence can replace the reward function.
    
    \item We perform a set of ablation studies across different sub-goal selection methods. We also evaluate HDT and the baselines on different data set types for the same task. The results of these studies can be found, along with the code at (place to be disclosed).
\end{enumerate}

\section{Related Work}
\label{sec:relatedwork}
Demonstration learning (DL) is a machine learning paradigm that allows robots to learn tasks from demonstrations performed by human experts and has been applied to multiple domains such as playing games, driving autonomously and robotics \cite{senhoras,surveyverboso,SchaalBillard,SchaalFuture,surveyosa,surveyzhu,recentadv}.
In this section, we provide a summary of some works related to ours.
The two predominant areas of demonstration learning are behavioral cloning (BC) and inverse reinforcement learning (IRL).

BC treats demonstration learning as a supervised learning problem, where the problem maps observations into actions, and the training signal is given by how similar the actions are to the demonstrator's \cite{mario}. 
On the other hand, IRL treats DL as a RL problem and DL is used to define the reward function used to estimate the policy \cite{shaping,shaping2,brys}. Both BC and IRL require multiple demonstrations and struggle to imitate tasks with long episodes.

Hierarchical approaches learn a high-level planner and a low-level controller \cite{iris,relay,swirl}.
The high-level planner finds a path built with sub-goal states that drive the agent towards the main task goal by conditioning the low-level controller to try to achieve each sub-goal. 
This extra guidance helps the agent learn in sparse reward environments and long tasks. 
Conditioning reinforcement learning and demonstration learning approaches on goal observations improves sample efficiency.

One way of improving sample efficiency is to learn an embedding space of skills from unstructured demonstrations \cite{accelerating} and then train the policy on the embedding space. Alternatively, complex tasks can be divided into smaller sub-tasks \cite{ltsd,swirl}. Each of these new sub-tasks are learnt from the demonstration data set and constitute sub-goals for the agent to reach. 
In \cite{ris} estimates a state distribution, to ensure the sub-goals are part reachable states.
Other approaches try to generalize new trajectories from the demonstration data using the insight that the trajectories intersect at certain states \cite{gti}.
A particularly promising approach was proposed, using goal conditioned policies at multiple layers of hierarchy for RL \cite{nachum,saenko}.
Alternatively, other methods generate sequences of sub-goals with a divide-and-conquer approach \cite{jurgenson,pertsch}.

Sequence modeling with deep networks has evolved from LSTMs to Transformer architectures with self-attention \cite{attention}. The latter have revolutionized many natural language processing tasks. 
Recently, they have been applied to RL by re-formulating it as a sequence modeling problem \cite{dt,tt}.
These treat reinforcement learning as a supervised learning paradigm that predicts action sequences from trajectories and task specification (e.g., target goal or returns), instead of traditionally learning Q-functions or policy gradients. A decision transformer \cite{dt} is a model-free context-conditioned policy, while the trajectory transformer is used both as a policy and model. The latter shows that beam search can be used to improve upon purely model-free performance. 
Later, \cite{odt} applied the original transformer to an online setting to fine tune the model using interaction data after pre-training on demonstration data, alleviating the limited performance inherited from the data set. In \cite{pretrain}, the authors showed that pre-training the decision transformer on large corpus of unrelated text data improves the performance of robotic tasks.

\section{Preliminaries}
\label{sec:preliminaries}

\subsection{Reinforcement Learning}

We consider learning in a Markov Decision Process (MDP) described by the tuple $(S, A, P, R)$. The tuple consists of states $s \in S$, actions $a \in A$, the state transition function $P(s'|s, a)$ and a reward function $r = R(s, a)$. We use $s_t$, $a_t$, and $r_t = R(s_t, a_t)$ to denote the state, action, and reward at timestep $t$, respectively. With sparse rewards, optimizing the expected discounted reward using RL may be difficult. Because of this, we augment this MDP with sets of absorbing goal and sub-goal states $G \subset S$ and $Sg \subset S$. Where each goal state $g \in G$ is a state of the world in which the task is considered to be solved and $sg \in Sg$ is a valuable state of the world which contributes to the success of the trajectory. A trajectory is a sequence of length $K$ of states, actions, and rewards: $\tau = (s_0, a_0, r_0, ... , s_K , a_K , r_K )$. The accumulated rewards of a trajectory $R_\tau$ with length $N$ are: $\sum_{t=1}^{N} r_t$. 
At every step $t$, the agent observes a state $s_t$ and queries the policy $\pi$ to choose an action $a_t = \pi(s_t)$.
The agent performs the action and observes the next state $s_{t+1} \sim P(s_{t+1}|s_t, a_t)$ and reward $r_t = R(s_t, a_t)$.
The goal is to learn a policy $\pi$ which produces trajectories $\tau$ which maximize the expected return $\mathop{\mathbb{E}}_\pi [R_\tau]$.

\subsection{Behaviour Cloning}

In behaviour cloning, instead of learning from experience by interacting with the environment, the agent learns solely from the trajectories present in the demonstration data set. This setting is harder because it removes the
ability for agents to explore the state space and try different trajectories. We assume there is access to a data set of trajectories $\mathop{\mathbb{D}} = \{\tau_1, ..., \tau_N\}$. In BC the agent is encouraged to select the action the demonstrator took for a given state in the data set. A common approach is to maximize the likelihood of actions in the demonstration, $\max\mathop{\mathbb{E}}_{(s,a)\sim \mathop{\mathbb{D}}}  log \pi(a|s)$.

\subsection{Decision Transformer}

Transformers are a group of architectures which model sequential data \cite{attention}. These are encoder-decoder models built with blocks of self-attention layers with residual connections. Due to their success in many natural language processing tasks, and because RL learns from sequences of trajectories, \cite{dt} created the decision transformer (DT). DT transforms RL into a sequence objective problem. Instead of processing a single state transition, the policy receives a sequence of 3 types of input tokens: $(rtg_1, s_1, a_1, ... , rtg_N, s_N, a_N)$. Corresponding to returns to go, states and actions. Where the value of the returns-to-go at timestep $t$ is the sum of future rewards in the trajectory $rtg_t = \sum_{t'=t}^{T} r_t'$. 

Additionally, the transformer receives a mask and the sequence of timesteps. The mask is a vector with as many elements as the length of the sequence. Each element in the mask indicates if the corresponding token should be hidden to the transformer. The sequence of timesteps is used for positional encoding of the tokens.

The policy is trained using the MSE loss, similarly to BC, to predict the demonstrator's action. DT feeds these trajectories through a GPT \cite{gpt} architecture. At test time, the agent doesn't have access to a demonstration trajectory to obtain the sequence of returns-to-go. Because of this, DT requires the user to specify the desired performance of the agent in the form of desired accumulated rewards. After each transition, we obtain the collected reward and decrement it from the specified desired accumulated reward value. The desired rewards can't be an arbitrarily high value, it must be a value within the limits of the accumulated rewards present in the demonstration data set used to train the agent. Additionally, in environments with sparse rewards, there can be large sequences where no reward is given. This results in cases where the values of the returns-to-go tokens are static in the sequence, which hinders the learning process of the agent. We aim to alleviate these two issues by replacing the sequence of returns to go with sequences of sub-goals specified by a high-level model.

\section{Proposed Approach}
\label{sec:proposed}

\begin{algorithm}[tb]
   \caption{HDT training algorithm.}
   \label{algorithm}
\begin{algorithmic}
    \STATE {\bfseries Input:} Demonstration data set $D = \{\tau_1,..., \tau_N\}$, batch size $B$, token size $K$, training iterations $I$, high-level mechanism $\pi_\phi$, low-level controller $\pi_\theta$, sub-goal selection method $f$.
    \FOR{$i=1$ {\bfseries to} $N$}
        \STATE $\tau'_i \leftarrow f(\tau_i)$
    \ENDFOR
    \FOR{$i=1$ {\bfseries to} $I$}
        \STATE \textbf{s}, \textbf{a}, \textbf{t}, \textbf{sg}, \textbf{mask} $\leftarrow$ Sample $B$ sub-trajectories from $D$.
        \STATE $sg' \leftarrow \pi_\phi(\textbf{s}, \textbf{sg}, \textbf{t}, \textbf{mask})$, obtain sub-goal predictions.
        \STATE $\phi \leftarrow L_\phi(sg, sg')$, update mechanism's weights.
        \STATE $a' \leftarrow \pi_\theta(\textbf{s}, \textbf{sg}, \textbf{a}, \textbf{t}, \textbf{mask})$, obtain action predictions.
        \STATE $\theta \leftarrow L_\theta(a, a')$, update controller's weights.
    \ENDFOR
    \STATE {\bfseries return} $\pi_\theta, \pi_\phi$
\end{algorithmic}
\end{algorithm}

\subsection{Overview}

We present the Hierarchical Decision Transformer (HDT), represented in Fig. \ref{hdt}. HDT is a hierarchical behaviour cloning algorithm which adapts the original decision transformer to tasks with longer episodes and/or sparse rewards, while also removing its dependency on desired returns specification. 
Here we first provide an overview of HDT and motivation of each component.
We split the decision making process of the agent into a high-level mechanism that sets sub-goal states for a low-level controller to try and reach.
We employ the decision transformer, which uses the GPT \cite{gpt} architecture, for both the high-level mechanism and the low-level controller. The high-level mechanism is conditioned on the state sequence and aims to produce sub-goal states for the low-level controller to reach, guiding it through the task. The low-level controller is similar to the original decision transformer, only it is conditioned on the sequence of sub-goals produced by the high-level model, instead of the sequence of returns-to-go.

The algorithm learns solely from demonstration data present in a data set. The data set is first processed, where for each state, a sub-goal state is selected from the trajectory using the sub-goal selection algorithm we present. The original transition is then augmented with the selected sub-goal state for training the models. Then, both the high-level mechanism and low-level controller are trained simultaneously using the sampled batches of sequences. 
The high-level mechanism receives sequences of the previous states and sub-goals and tries to predict the next sub-goal state in the sequence. The low-level controller receives sequences of the previous states, sub-goals and actions, and tries to predict the next action in the sequence.
Because both models are transformers, they also receive the sequence of time-steps and the mask for positional encoding and to hide the future tokens from the decoder, respectively.

The complete training loop is provided in Algorithm \ref{algorithm}.
We encourage training both models simultaneously to avoid repeating steps, such as batch sampling. However, because both models don't rely on data produced by the other to learn, they can be trained sequentially.
Next, we further explain each component.

\subsection{Sub-Goal Selection}
\label{sec:proposed_selection}

Our method requires sequences of states, actions, time-steps and sub-goals. To obtain these sequences, it processes the demonstration data set before training. The sequences of states, actions and time-steps can be directly extracted from the data set. However, the sub-goal sequences aren't explicitly present in the data set and must be inferred. We define sub-goals $sg$ in regards to the current state $s_t$ as later states in the trajectory, which are highly valuable for the agent to reach. These states should mark milestones in the trajectory which when reached sequentially, would make it highly probable that the agent successfully performs the task.

Using this definition, the sub-goals guide the agent through the task, meaning they have the same purpose as the returns-to-go. Therefore, we replace the returns-to-go sequences with the sub-goal sequences. This way, the user doesn't need to iterate over the value of the desired returns for each task in order for the method to perform. Additionally, it replaces null rewards in tasks with sparse rewards, consequently improving the learning process. We perform experiments to show that the original decision transformer is dependent on the returns-to-go sequences to learn and that the value of the desired returns is important.

By our definition, a sub-goal state $sg$ has high value for the success of the trajectory. We can model this behaviour by finding a following state $s_j$ in the trajectory with high values of accumulated rewards, since the current state $W(s_j) = \sum_{k=i+1}^{j} r_k$. However, this would always prioritize the last states of the trajectory. To encourage selecting states close to the current state, we divide the accumulated rewards by the distance between the states: $W(s_j) = \sum_{k=i+1}^{j} \frac{r_k}{j-i}$.

The weighted average of the accumulated rewards in a sub-trajectory identifies sub-goal states that have correctly completed a part of the task resulting in a significant reward, without always selecting far away states, by punishing the length of the sub-trajectory. We select the state with highest associated weight to be the sub-goal.

\subsection{Low-Level Controller}

The low-level controller is a goal conditioned Decision Transformer $\pi_{\theta}$. Similar to decision transformer \cite{dt}, we model the demonstration trajectories as sequences of tokens for the transformer to learn from. The low-level controller receives sequences of states, actions and sub-goals and tries to predict the next action.
During training, sequences of size $K$ are randomly sampled in a batch from the enhanced data set, where $K$ indicates the number of tokens. The low-level transformer is trained to minimize the Behavioral Cloning loss:

$\mathop{\mathbb{L}}_\theta (s_{t:t+K}, a_{t:t+K}, sg_{t:t+K}) = \|a_{t:t+K} - \pi_{\theta} (s_{t:t+K}, a_{t:t+K}, sg_{t:t+K})\|$.

For each sequence, the model also receives the respective token time-steps for positional encoding and a mask to hide the future tokens from the decoder. We don't include them in the formulas for simplicity.
At test time, the sequences are built by keeping a history of past states, actions and time-steps from the environment and past sub-goals selected from the high-level mechanism.

\subsection{High-Level Mechanism}
The high-level mechanism chooses sub-goal states for the low-level controller to try and reach. It is a state conditioned Decision Transformer $\pi_{\phi}$. We use the GPT architecture \cite{gpt} for both high and low-level transformers. The high-level controller receives the sequence of states, past sub-goals, time-steps and mask, and aims to predict the next sub-goal state which will help the low-level controller succeed in the task. During training, the high-level mechanism is trained to minimize its own Behavioural Cloning loss:

$\mathop{\mathbb{L}}_\phi (s_{t:t+K}, sg_{t:t+K}) = \|sg_{t:t+K} - \pi_{\phi} (s_{t:t+K}, sg_{t:t+K})\|$.

\section{Experiments}
\label{sec:experiments}

In this section, we evaluate the performance of the Hierarchical Decision Transformer and compare it with the original DT and BC baselines.
We evaluate the models on a varied subset of tasks from OpenAIGym \cite{openai}, D4RL \cite{d4rl} and RoboMimic \cite{robomimic} benchmarks. The tasks range from short episodes to long episodes and from frequent rewards to sparse reward settings. We train the algorithms on the different data sets provided by the benchmarks for each task.
For DT, we first find the maximum accumulated returns collected by a trajectory in the demonstration data set. We then set the desired returns to this value and also to half, following the original tests of DT. 
We use the decision transformer architecture for both low and high level models.
We train each model for 100 thousand epochs, using batch sizes of 64, learning rate of $1e^{-4}$ and sequence length of 20 tokens.
To reduce the impact of outliers and because the episodes are seed dependent, every one thousand epochs we validate the model on 100 episodes and calculate the average accumulated rewards.
In the tables, we present the highest average accumulated rewards seen throughout the 100 thousand epochs.

\subsection{Does DT rely on desired returns?}

\begin{table}[tb]
\caption{Maximum accumulated returns of the original DT and of a DT variant without the desired returns input sequence trained for 100 thousand iterations.}
\label{tab:dtrvsnor}
\centering
\begin{tabular}{ccccc}
\hline
\textbf{Task}                 & \textbf{Data Set}         & \textbf{\begin{tabular}[c]{@{}c@{}}Desired\\ Return\end{tabular}} & \textbf{\begin{tabular}[c]{@{}c@{}}Original\\ DT\end{tabular}} & \textbf{\begin{tabular}[c]{@{}c@{}}DT w/o\\ Desired Returns\end{tabular}} \\ \hline
\multirow{3}{*}{Half-Cheetah} & \multirow{3}{*}{medium}   & 2655                                                              & \textbf{5095}                                                    & 47                                                                        \\
                              &                           & 5309                                                              & \textbf{5092}                                                  & 47                                                                          \\
                              &                           & 10000                                                             & \textbf{4953}                                                   & 47                                                                           \\ \hline
\multirow{3}{*}{Hopper}       & \multirow{3}{*}{medium}   & 1611                                                              & \textbf{2303}                                                & 300                                                                        \\
                              &                           & 3222                                                              & \textbf{2557}                                                  & 300                                                                     \\
                              &                           & 10000                                                             & \textbf{2036}                                                  & 300                                                                      \\ \hline
\multirow{3}{*}{Kitchen}      & \multirow{3}{*}{complete} & 2                                                                 & \textbf{2.4}                                                   & 1.0                                                                     \\
                              &                           & 4                                                                 & \textbf{2.5}                                                   & 1.0                                                                      \\
                              &                           & 10000                                                             & \textbf{2.1}                                                   & 1.0                                                                      \\ \hline
\multirow{3}{*}{Walker-2D}    & \multirow{3}{*}{medium}   & 2113                                                              & \textbf{3619}                                                  & 149                                                                      \\
                              &                           & 4227                                                              & \textbf{3711}                                                  & 149                                                                      \\
                              &                           & 10000                                                             & \textbf{3100}                                                  & 149                                                                      \\ \hline
\end{tabular}
\end{table}

One of the motivations of our work is to remove the requirement of specifying the value of the desired returns to the Decision Transformer. We first need to know whether the DT truly depends on the desired returns sequence to perform or if they can simply be removed from the model's input. Table \ref{tab:dtrvsnor} shows the accumulated returns obtained by the original DT model and by a DT model variant without the desired returns sequence. For the original DT, we set value of the desired returns for the task to the maximum value of accumulated rewards present in the data set and to half of this value. 
We also set the value of the desired return to an arbitrarily high value, for example 10000.

The results in Table. \ref{tab:dtrvsnor} show that the DT model is dependent on the desired returns to perform. Removing them prevents DT from learning the task. Additionally, it also shows that the value of the desired returns isn't obvious. Because DT only reaches the desired return of half of the maximum value on the kitchen task. Then, when doubling the value of the desired returns, it only slightly increases the performance. Lastly, when using an arbitrarily high value, DT under-performs. In conclusion, 
the value of the desired returns is crucial for the performance of the DT model while being non-trivial to determine.

\subsection{Does HDT replace the need for desired returns?}

\begin{table}[tb]
\caption{Maximum accumulated returns of the HDT and of a HDT variant including the desired returns input sequence trained for 100 thousand iterations.  }
\label{tab:oursrvsnor}
\centering
\begin{tabular}{ccccc}
\hline
\textbf{Task}                 & \textbf{Data Set}         & \textbf{\begin{tabular}[c]{@{}c@{}}Desired\\ Return\end{tabular}} & \textbf{HDT} & \textbf{\begin{tabular}[c]{@{}c@{}}HDT with \\ Desired Returns\end{tabular}} \\ \hline
\multirow{2}{*}{Half-Cheetah} & \multirow{2}{*}{medium}   & 2655 & \textbf{5205} & 5004 \\
                              &                           & 5309 & \textbf{5205} & 5002 \\
\multirow{2}{*}{Hopper}       & \multirow{2}{*}{medium}   & 1611 & \textbf{3071} & 1213 \\
                              &                           & 3222 & \textbf{3071} & 1198 \\
\multirow{2}{*}{Kitchen}      & \multirow{2}{*}{complete} & 2 & \textbf{2.6}  & 2.1 \\
                              &                           & 4 & \textbf{2.6}  & 1.4 \\
\multirow{2}{*}{Walker 2D}    & \multirow{2}{*}{medium}   & 2113 & \textbf{3879} & 3645 \\
                              &                           & 4227 & \textbf{3879} & 3516 \\ \hline
\end{tabular}
\end{table}

Next, we test whether the high-level mechanism, by selecting sub-goals, replaces the need for the sequence of desired returns, in regards to guiding the agent through the task. We test this hypothesis by adding the sequences of desired returns as an extra input to the low-level controller and compare the performance with our base method without this extra input. Table \ref{tab:oursrvsnor} shows the accumulated returns obtained by our base HDT model and by the HDT model variant with the extra sequence of desired returns. 

The results show that HDT performs slightly better without the desired returns sequence. This means that the high-level mechanism is able to output sub-goal states from the history of past states, capable of guiding the low-level controller through the task and consequently replacing the need for desired returns. Consequently, it becomes possible to remove the requirement for external knowledge about the specific task.

\subsection{Baseline Comparison}

We compare HDT with DT and BC baselines and present the results in Table \ref{tab:full}. 
For BC, we used an MLP with two fully connected layers. We include the average accumulated rewards collected by the demonstrator in the data set. 
Additionally, the table shows the average length of the trajectories present in the data set.
HDT outperforms the baselines in eight out of ten tasks. This means that our method successfully improved upon the original DT in raw performance while also removing its dependency on desired reward specification, making it task-independent. Particularly, the Maze 2D constitutes a task with sparse rewards where, on average 90\% of the transitions receive no reward, and our method vastly outperforms the baselines. During the execution of the kitchen task, the agent only receives a reward after completing each of the four sub-tasks making it a long-horizon task. The discrepancy between the transformer methods and BC is more prominent in these types of tasks. We conclude that our method offers superior performance overall, and is more robust to both long-horizon and sparse reward tasks.

\begin{table}[tb]
\caption{Maximum accumulated returns of the HDT compared to DT and BC baselines.  }
\label{tab:full}
\centering
\begin{tabular}{ccccccc}
\hline
\textbf{Task} & \textbf{Data Set} & \textbf{AVG Length} & \textbf{AVG Rewards} & \textbf{HDT}  & \textbf{DT}  & \textbf{BC}  \\ \hline
Ant           & medium            & 832                 & 3051                 & \textbf{3993} & 3943         & 3952         \\
Door          & human             & 269                 & 796                  & 338           & \textbf{459} & 363          \\
Half-Cheetah  & medium            & 1000                & 4770                 & \textbf{5205} & 5095        & 5198         \\
Hammer        & human             & 452                 & 3072                 & \textbf{3115} & -68          & 50           \\
Hopper        & medium            & 457                 & 1422                 & \textbf{3071} & 2557         & 1969         \\
Kitchen       & complete          & 194                 & 325                  & \textbf{2.6}  & 2.5          & 0.8          \\
Maze 2D       & medium            & 168                 & 4                    & \textbf{188}  & 68           & 37           \\
Pen           & human             & 200                 & 6326                 & \textbf{3897} & 3606         & 867          \\
Relocate      & human             & 398                 & 3674                 & 35            & 35           & \textbf{232} \\
Walker 2D     & medium            & 840                 & 2852                 & \textbf{3879} & 3711         & 3723         \\ \hline
\end{tabular}
\end{table}

\section{Conclusions}
\label{sec:conclusions}

We present the Hierarchical Decision Transformer (HDT), where we adapt the original Decision Transformer (DT) into an hierarchical setting. By doing so, we improve its performance in regards to accumulated rewards overall. Particularly, HDT is more robust to tasks with longer episodes and less sensitive to environments with sparse rewards. Moreover, we remove the requirement for external information about the task in regards to the user having to specify desired returns beforehand. 

Lastly, we define a sub-goal selection method which does not require any task knowledge which makes our method fully task-independent.
Future work will aim at evaluating further the HDT's performance on more tasks and using different architectures for both models.

\end{document}